\documentclass[format=sigconf]{acmart}

\AtBeginDocument{%
  \providecommand\BibTeX{{%
    \normalfont B\kern-0.5em{\scshape i\kern-0.25em b}\kern-0.8em\TeX}}}

\setcopyright{acmcopyright}
\copyrightyear{2018}
\acmYear{2018}
\acmDOI{10.1145/1122445.1122456}

\usepackage{array}
\usepackage{multirow}
\usepackage{amsmath}
\begin{document}

\title{Automatic Generation of Product-Image Sequence in E-commerce}

\author{Xiaochuan Fan}
\orcid{}
\affiliation{%
  \institution{JD.COM Research}
  \streetaddress{}
  \city{Mountain View}
  \country{USA}
  \postcode{}
}
\email{efan3000@gmail.com}

\author{Chi Zhang}
\affiliation{%
  \institution{JD.COM Research}
  \streetaddress{}
  \city{Mountain View}
  \country{USA}}
\email{zhangc11msu@gmail.com}

\author{Yong Yang}
\affiliation{%
  \institution{JD.COM}
  \streetaddress{}
  \city{Beijing}
  \country{China}}
\email{yangyong280@jd.com}

\author{Yue Shang}
\affiliation{%
  \institution{JD.COM Research}
  \streetaddress{}
  \city{Mountain View}
  \country{USA}}
\email{shangyue1230@gmail.com}

\author{Xueying Zhang}
\affiliation{%
  \institution{JD.COM Research}
  \streetaddress{}
  \city{Mountain View}
  \country{USA}}
\email{zxycontact@gmail.com}

\author{Zhen He}
\affiliation{%
  \institution{JD.COM}
  \streetaddress{}
  \city{Beijing}
  \country{China}}
\email{bjhezhen@jd.com}

\author{Yun Xiao}
\affiliation{%
  \institution{JD.COM Research}
  \streetaddress{}
  \city{Mountain View}
  \country{USA}}
\email{yun.xiao@yahoo.com}

\author{Bo Long}
\affiliation{%
  \institution{JD.COM}
  \streetaddress{}
  \city{Beijing}
  \country{China}}
\email{bo.long@jd.com}

\author{Lingfei Wu}
\affiliation{%
  \institution{JD.COM Research}
  \streetaddress{}
  \city{Mountain View}
  \country{USA}}
\email{lwu@email.wm.edu}

\renewcommand{\shortauthors}{Xiaochuan Fan et al.}

\begin{abstract}
Product images are essential for providing desirable user experience in an e-commerce platform. For a platform with billions of products, it is extremely time-costly and labor-expensive to manually pick and organize qualified images. Furthermore, there are the numerous and complicated image rules that a product image needs to comply in order to be generated/selected. To address these challenges, in this paper, we present a new learning framework in order to achieve Automatic Generation of Product-Image Sequence (AGPIS) in e-commerce. To this end, we propose a Multi-modality Unified Image-sequence Classifier (MUIsC), which is able to simultaneously detect all categories of rule violations through learning. MUIsC leverages textual review feedback as the additional training target and utilizes product textual description to provide extra semantic information. 
Based on offline evaluations, we show that the proposed MUIsC significantly outperforms various baselines. Besides MUIsC, we also integrate some other important modules in the proposed framework, such as primary image selection, non-compliant content detection, and image deduplication. With all these modules, our framework works effectively and efficiently in JD.com recommendation platform. By Dec 2021, our AGPIS framework has generated high-standard images for about 1.5 million products and achieves 13.6\% in reject rate. Code of this work is available at https://github.com/efan3000/muisc.

\end{abstract}

%



\copyrightyear{2022}
\acmYear{2022}
\setcopyright{acmcopyright}\acmConference[KDD '22]{Proceedings of the 28th ACM SIGKDD Conference on Knowledge Discovery and Data Mining}{August 14--18, 2022}{Washington, DC, USA}
\acmBooktitle{Proceedings of the 28th ACM SIGKDD Conference on Knowledge Discovery and Data Mining (KDD '22), August 14--18, 2022, Washington, DC, USA}
\acmPrice{15.00}
\acmDOI{10.1145/3534678.3539149}
\acmISBN{978-1-4503-9385-0/22/08}

\maketitle

\section{Introduction} \label{introduction}
\begin{figure}[t!]
  \includegraphics[width=0.45\textwidth]{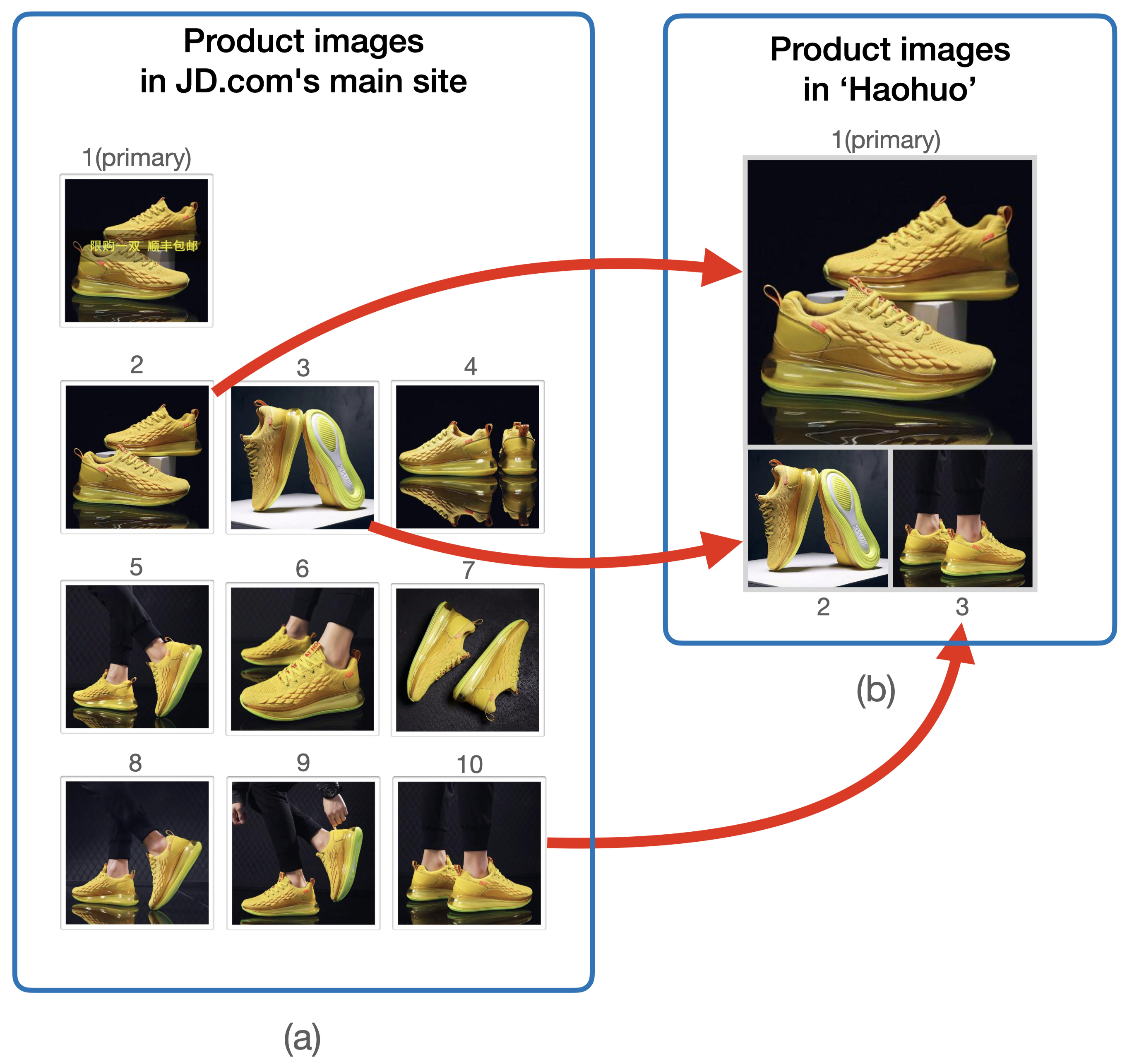}
  \caption{A pair of shoes are presented by images from (a) JD.com's main site and (b) JD.com's "Haohuo" Channel. (a) contains redundant and non-compliant content, while (b) presents all aspects of the product via three images.}
  \Description{}
  \label{fig:main_vs_haohuo}
\end{figure}

Product images are essential for the success of online shopping \cite{chen2013comprehensive}\cite{di2014picture}\cite{zakrewsky2016item}, as good product images enhance product description and help fill the gap between offline shopping and online shopping. Most of well-known e-commerce platforms have their own various complicated standards on product images, such as Amazon, Ebay, Alibaba and JD.com. "Haohuo" channel\footnote{https://fxhh.jd.com/} (referring to "Discovery Goods"), an important
traffic entrance on the top of JD.com (both website and App), is featured by its high standard of product presenting. A typical process for selecting product images are following this procedure: i) "HaoHuo" channel motivates human professions to submit high-quality product images; ii) human reviewers then review the submitted images according to a set of complicated image standards, which are largely designed based on in-house experts' experience and A/B tests. Figure \ref{fig:main_vs_haohuo} exhibits an example of comparison between product images from JD.com's main site and "Haohuo" channel. We can see that "Haohuo" channel presents the product in a concise and well-organized way. However, for an e-commence platform with large-scaled product catalogs, it is extremely time-costly and labor-expensive to manually pick product images. In this paper, we propose a novel learning framework for Automatic Generation of Product-Image Sequence (AGPIS), which automatically picks a sequence of product images from candidate images according to a set of numerous and complicated rules. 

\begin{figure*}[t]
  \includegraphics[width=1.0\textwidth]{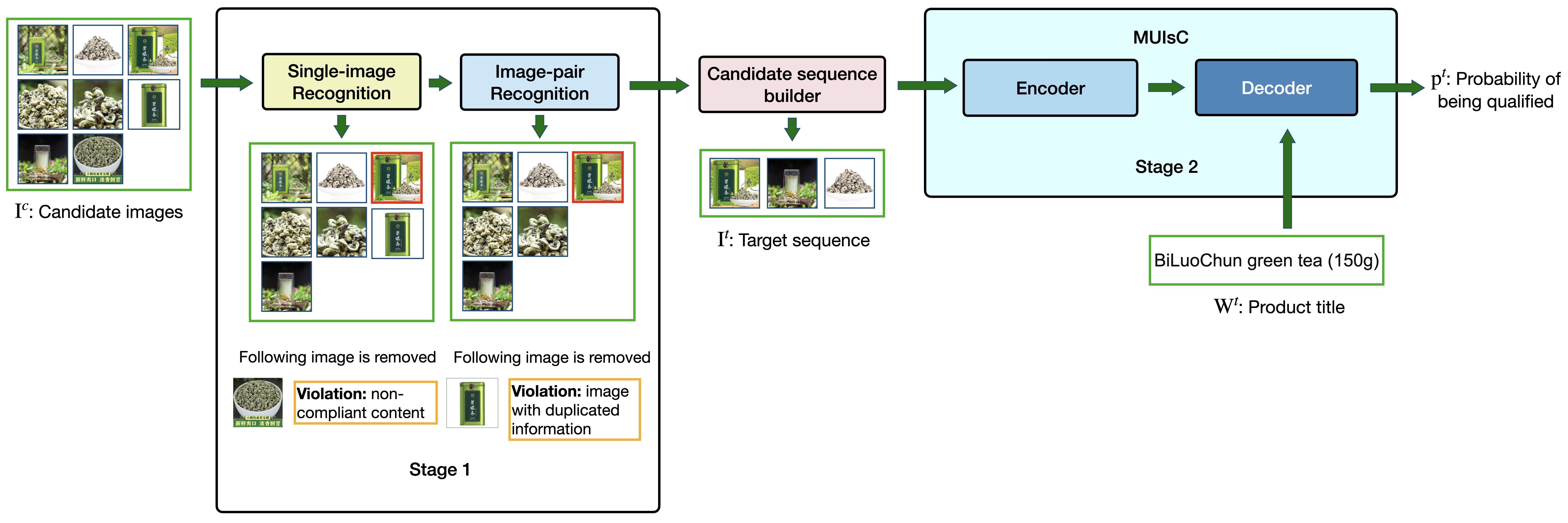}
  \caption{Pipeline of our AGPIS framework. Stage 1 detects  non-compliant/redundant images and selects the primary image as shown by the red image border. Given a target image sequence built from stage 1, stage 2 outputs its probability of being qualified.}
  \Description{}
  \label{fig:pipeline}
\end{figure*}

There are some early works to address similar problems \cite{chaudhuri2018smart}\cite{gandhi2019image}, which however are inadequate to address the aforementioned problems due to various unique challenges. Firstly, there are numerous and complicated rules and these rules could change over the time. This makes it unaffordable to develop rule-based specific methods or datasets for each rule due to the huge computation as well as the development and maintenance costs. Furthermore, the existing methods only focus  on one or several rules \cite{chaudhuri2018smart}\cite{gandhi2019image}. As a result, it is hard to work well with these systems since they are designed and customized with the fixed prior knowledge of rules. For example, detection of non-compliant content often involves logo detection \cite{joly2009logo}\cite{romberg2013bundle} and skin region detection \cite{yin2011big}\cite{ jones2002statistical}. 

Secondly, different rules may require different information. Product image rules can be generally divided into three categories as follows. First, \textbf{Single-image rules.} Typical single-image rules are about image quality, e.g. unnatural artifact or blurry image, and  non-compliant content, e.g. logo, banner, and water-print. The detection of single-image rule violations is only relevant to an individual image. 
Second, \textbf{Image-pair rules.} Image-pair rules involve matching and comparison of two images in order to avoid redundant or wrong information. An example of image-pair rules states that two images should not present product appearance from similar viewing angles. The detection of this category of violations requires information from a pair of images.
Third, \textbf{Multi-image rules.} These rules are usually designed for the layout of product images to make sure product information is adequate and released in a proper order. The detection of multi-image rule violations may require information from multiple images or even cross-modality product description. Most of the existing methods \cite{gandhi2019image}\cite{joly2009logo}\cite{romberg2013bundle}\cite{yin2011big}\cite{ jones2002statistical} only focus on violation detection of single-image and image-pair rules, and ignore the exploration of relations between all the images in a sequence. However, such relations are essential for automatic image selections since a product is presented by multiple images as a whole. 

Last but not least, the rich information of image review feedback can be used for automatic image selection. The existing methods usually cast their problems into image classification problems. Then these product images are labeled as either qualified or not qualified according to the rules. However, textual feedback from human reviewers may contain rich semantic information which cannot be easily converted to classification labels. Taking "Haohuo" channel as an example, besides the violated rule name, a review feedback may also include extra information to explain these rejections, including which image in a sequence violates the rule, the location of non-compliant content in an image or what the non-compliant content is. Therefore, such rich semantic information is also helpful for improving automatic image selection. 

To address these aforementioned challenges, in this paper, we present a novel learning framework in order to achieve Automatic Generation of Product-Image Sequence (AGPIS) in e-commerce. The core module of our framework is a Multi-modality Unified Image-sequence Classifier (MUIsC), which is able to simultaneously detect all categories of rule violations through learning. Firstly, to obtain adequate information for AGPIS, MUIsC takes as input an image sequence, rather than a single image or a pair of images, and extracts features via a hierarchical encoder. Then, along with the classification task, we use Natural Language Generation (NLG) of image review feedback as an auxiliary training task to fully exploit the rich semantic information of a review feedback. Different from some traditional tasks such as visual question answering \cite{antol2015vqa} and image captioning \cite{vinyals2015show,liu2021cptr}, which aim to generate text conditioned on a visual input, our NLG task functions as a guide for MUIsC to better understand complicated rules during training. 
In addition, the introduction of NLG task does not incur any additional burden in data processing since no manual labeling is needed. This is especially important for frequent model update in a real application. Lastly, textual product description is also fed to MUIsC as a input to assist image recognition. This requires our proposed MUIsC to be able to perform image-text interaction effectively. Interestingly, 
the resulting MUIsC has similar input and output with a human reviewer, which means that no prior knowledge or rule-based task is involved in our model. 

In order to accumulate data for MUIsC training and make our framework work efficiently, we also integrate additional modules for single-image and image-pair recognition to detect unqualified images and build a sequence candidate for MUIsC. Figure \ref{fig:pipeline} exhibits the overall pipeline of the proposed framework. Given a set of candidate images and textual product description, our framework outputs an image sequence and its probability of being qualified. If the probability is larger than a threshold, we submit the image sequence to a human reviewer and then "Haohuo" channel if approved. The red arrows in Figure \ref{fig:main_vs_haohuo} indicates an example of the correspondence between candidate images (from JD's main site) and the resulting image sequence of our framework for "Haohuo" channel.

Since its deployment in JD.com in Feb 2021, our AGPIS framework has generated high-standard images for about 1.5 million products and achieves 13.6\% in reject rate.

\section{Related Works}
In this section, we introduce works on similar topics and related domains.

\textbf{E-commerce image selection}.
The significance of images in e-commerce has been well-studied \cite{chen2013comprehensive}\cite{di2014picture}\cite{zakrewsky2016item}, but e-commerce retailers that offer marketplace platforms still struggle to control image quality. 
Gandhi et al. \cite{gandhi2019image} resolves the issue of non-compliant content by combining state-of-the-art image classification and object detection models, but non-compliant content is only part of numerous and complicated rules for automatic image selection.
Chaudhuri et al. \cite{chaudhuri2018smart} proposes a system that aggregates images from various suppliers to produce a image set, arranged in an order according to a set of manually-crafted templates. The template designer has to design different templates for different product categories. This costs huge human efforts and leads to low production efficiency. More importantly, this method limits the diversity of visual product presenting, i.e. the products in the same category are presented in a similar style. By contrast, our proposed framework learns to organize images from a large-scaled reviewed data, which is much more flexible. Besides, in \cite{gandhi2019image} and \cite{chaudhuri2018smart}, systems consist of a sequence of modules designed for every specific rule, while our single MUIsC model is able to detect all categories of rule violations.

\textbf{Image aesthetic and quality assessment} has received considerable attention in recent years. This technique, which is similar to violation detection of single-image rules in automatic image selection,  has been widely used in user album photo selection \cite{kuzovkin2019context} and image recommendation \cite{yu2018aesthetic}.
For conventional image quality assessment methods, hand-craft features created from either photography practices or objective quality criteria are widely used \cite{datta2006studying}\cite{ke2006design}\cite{mavridaki2015comprehensive}. 
More recently, learnt feature representations using deep neural network has surpassed the performance of hand-craft  ones \cite{kang2014convolutional}\cite{jin2016image}\cite{sheng2020revisiting}. Most of the image assessment works are reported on datasets with assessment scores from peer reviewers, and cast the problem into image classification or ranking. Comparatively, textual product description and review feedback are in the form of natural language and have much richer semantic information than numeric scores. Besides, our framework confronts more complicated image rules and handles image sequences rather than single images.

\textbf{Vision-and-language (VL) models}, leveraging information from both modalities, has been a very active topic recently. Since transformer\cite{vaswani2017attention} based models are adopted in computer vision, VL models achieved great success in tasks such as Visual Question Answering (VQA) \cite{antol2015vqa}, image captioning \cite{vinyals2015show}\cite{liu2021cptr}, and image-text matching \cite{lee2018stacked}, etc. Most of these works follow the model architecture in variants of visual backbones (ResNet\cite{he2016deep}, ViT\cite{dosovitskiy2020image}, CLIP\cite{radford2021learning}), text encoder/decoder (Bert\cite{devlin2018bert}, Roberta\cite{liu2019roberta}, GPT\cite{radford2019language}), modality-fusion scheme (single-stream\cite{li2019visualbert}, multi-stream\cite{lu2019vilbert}\cite{hu2021unit}), and pre-training objectives (masked language modeling\cite{devlin2018bert}, masked image modeling \cite{chen2020uniter}\cite{tan2019lxmert}, multimodal alignment\cite{lu2019vilbert,radford2021learning,tan2019lxmert}). 
Our MUIsC model also follows the transformer based text-image modeling framework.  Specifically we use a transformer-based encoder for visual feature extraction and adopt an encoder-decoder (two-stream) architecture\cite{vaswani2017attention} to fuse the visual and text information encoded by separate encoders. Pre-trained visual and language models are also used for model parameter initialization.

\section{Method}

\subsection{Data and Notations}\label{problem}
In this paper, we propose a framework for Automatic Generation of Product-Image Sequence (AGPIS), which picks a sequence of images from a set of candidate images according to a set of rules. Our work requires data from two different sources, JD.com's main site and JD.com's "HaoHuo" channel. From JD.com's main site, we collect product images as candidates and product title as a textual product description. From "HaoHuo" channel, we collect reviewed image sequences and corresponding textual feedback for model training and evaluation.

We adopt the following notations for our data. A set of $K^c$ candidate images are represented by $\mathbf{I}^c = \{{\mathbf{i}}^c_1,...,{\mathbf{i}}^c_{K^c}\}$, where ${\mathbf{i}} \in {\mathbb{R}^{H{\times}W{\times}3}}$ denotes a RGB image, $H$ and $W$ are height and width of an image. Similarly, an image-sequence that we aim to generate, called target image sequence, is represented by $\mathbf{I}^t = \{ {\mathbf{i}}^t_1,...{\mathbf{i}}^t_{K^t}\}$, where $\mathbf{I}^t\in\mathbf{I}^c$ and $K^t\leq{K}^c$ is the number of images in $\mathbf{I}^t$. Taking the dataset used in this work as an example, a product in JD.com's main site has about 7 images on average, i.e. ${K^c}\approx7$, and a target sequence consists of 3 images, i.e. ${K^t}=3$. Note that our framework can be also applied to the problem with various values of ${K^t}$ and ${K^c}$. Textual feedback and product title can both be represented by a sequence of words, $\mathbf{W}^f = \{\mathbf{w}^f_1,..., \mathbf{w}^f_{K^f}\}$ and $\mathbf{W}^t = \{ \mathbf{w}^t_1,...,\mathbf{w}^t_{K^t}\}$, respectively. For accepted image sequences, we set their textual feedback to a constant word, e.g. "yes", i.e. $K^f = 1$ and $w^f_1 = \textrm{"yes"}$.

\subsection{Framework Overview}
Figure \ref{fig:pipeline} provides the overview of our proposed framework for automatic image selection. The framework consists of two stages. Stage 1 consists of a single-image recognition module and an image-pair recognition module. Given a set of candidate images $\mathbf{I}^c$, the single-image recognition module selects the primary (first) image for the target sequence and detects non-compliant content. Following the single-image recognition module, the image-pair recognition module detects the violations of image-pair rules. Then, a target image sequence $\mathbf{I}^{t}$ is built from the remaining images in $\mathbf{I}^c$. In stage 2, $\mathbf{I}^t$ and its corresponding textual description $\mathbf{W}^t$ are fed into a model named Multi-modality Unified Image-sequence Classifier (MUIsC), which estimates the probability of $\mathbf{I}^t$ being qualified, denoted as $p^t$. $\mathbf{I}^t$ and $p^t$ are the final output of our framework. If $p^t$ is larger than a threshold $th^t$, we 
send $\mathbf{I}^{t}$ to a human reviewer and then "Haohuo" channel if approved.

\subsection{Stage 1}
\subsubsection{Single-image Recognition}
The single-image recognition module consists of two learning-based models. One selects the primary image for $\mathbf{I}^{t}$ since some rules are specifically designed for primary image, and the other model detects non-compliant images. Each of these two models works as a binary single-image classifier based on Deep Neural Networks (DNNs). 

For the learning of primary-image selection model, we collect data from image sequences approved by human reviewers ${\mathbf{I}^{t*}}$ and their candidate images $\mathbf{I}^{c*}$. We label an image, denoted as ${\mathbf{i}} \in {\mathbf{I}^{t*} \cup \mathbf{I}}^{c*}$, to $j^{primary}$ as follows,
\begin{equation} 
j^{primary} = \begin{cases} 1 &  {\textbf{i}}={\textbf{i}}^{t*}_1 \\
                     0 & \textrm{otherwise}
     \end{cases}
\end{equation}, where ${\mathbf{i}}^{t*}_1$ is the primary image of $\mathbf{I}^{t*}$.

We train the model on a binary classification task using a cross-entropy
loss function. Let the model's output be $p^{primary}$
, which is the
probability that ${\mathbf{i}}$ is a primary image. During inference, the image with the largest $p^{primary}$ in $\mathbf{I}^c$ is selected as the primary image, and $p^{primary}$ should be larger than a threshold. If no such image exists, the process of AGPIS is terminated and the framework outputs nothing.

Our model for non-compliant image detection has similar architecture and loss function with the primary-image selection model but uses a different dataset. From reviewed image sequences, we picked the ones which are rejected due to non-compliant content. Then the images in these sequences are manually labeled into two groups: compliance and non-compliance. Let the model's output for non-compliant image detection be $p^{nc}$
, which is the
probability that an image contains non-compliant content. All images with $p^{nc}$ larger than a threshold are removed from $\mathbf{I}^c$ in the inference stage. If the number of remaining images is less than ${K^t}$, the process of AGPIS is terminated and the framework outputs nothing. 

\begin{figure}[t!]
  \includegraphics[width=0.35\textwidth]{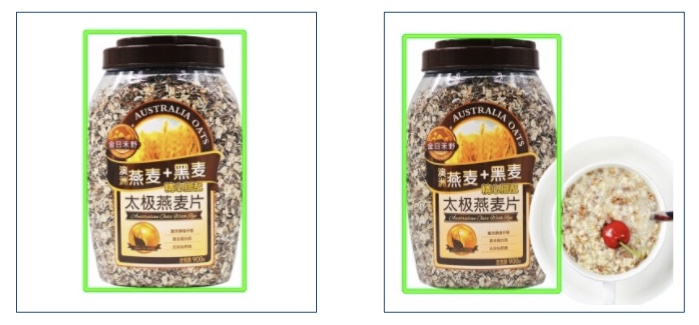}
  \caption{An example image pair that violates the rule "images with duplicated content". The green boxes show the duplicated patch-pair detected by our method.}
  \Description{}
  \label{fig:duplicated_pair_sample}
\end{figure}

\begin{figure*}[t]
  \includegraphics[width=0.99\textwidth]{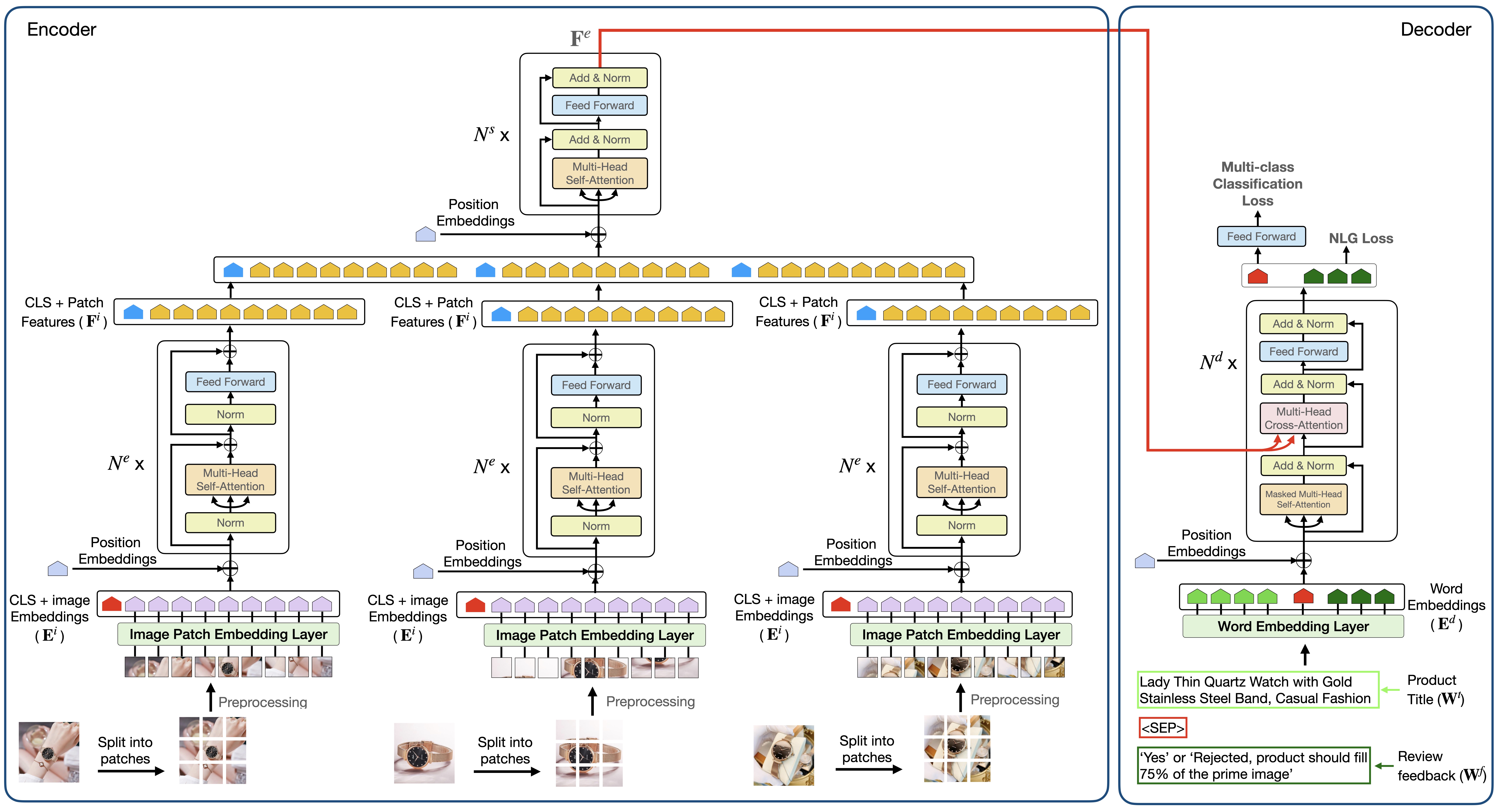}
  \caption{Architecture of MUIsC}
  \Description{}
  \label{fig:unified_model_big}
\end{figure*}

\subsubsection{Image-pair Recognition}
The module of image-pair recognition detects the violations of image-pair rules. In the real applications, this module is required to match and compare a pair of image patches rather than the whole images. Figure \ref{fig:duplicated_pair_sample} shows a pair of images which violate one of our image-pair rules. We can see that these two images contain much duplicated visual information although they do not have exactly the same content. Therefore, a main difficulty in detection of this category of violations is the construction of a set of region proposals which contain meaningful product information. The representative region proposal algorithms include two-stage object detectors (e.g. Faster RCNN \cite{ren2015faster}), Selective Search \cite{uijlings2013selective} and EdgeBoxes \cite{zitnick2014edge}. We empirically find that two-stage object detectors perform badly on e-commerce product images, and speculate that this is caused by different data distribution. In our method, we adopt EdgeBoxes to produce a certain number of region proposals for each image in $\mathbf{I}^c$, and use a pretrained DNN to extract patch features. Then patches from two images are matched and compared based on k nearest neighbor algorithm. In Figure \ref{fig:duplicated_pair_sample}, a pair of detected duplicated patches are outlined by green boxes. If a violation is detected in an image pair, the image with 
smaller $p^{nc}$ or the selected primary image is kept while the other image is removed. If the number of remaining images is less than ${K^t}$, our framework outputs nothing. 
\\
\\
Based on the output images of stage 1, a target sequence $\mathbf{I}^t$ is built. The selected primary image is the first image of $\mathbf{I}^t$. Then, we randomly selected $K^t-1$ images from the remaining images.

\subsection{Stage 2: MUIsC}

Our Multi-modality Unified Image-sequence Classifier (MUIsC) is designed following the transformer encoder–decoder architecture illustrated in Figure \ref{fig:unified_model_big}. The encoder extracts features for a given image sequence using a hierarchical architecture, while the decoder performs vision-language fusion, and estimates textual feedback and classification probabilities.

\subsubsection{Encoder}
Given ${\mathbf{I}^t}$ built after stage 1, the encoder firstly uses a Vision Transformer (ViT) \cite{dosovitskiy2020image} to extract feature for each image ${\mathbf{i}}^t \in {\mathbb{R}^{H{\times}W{\times}3}}$  separately. A ViT has two main steps: image patch embedding generation and transformer encoding. In the first step, an image is split into $N^p$ fixed-sized square patches, where $N^p=(W/P)\times(H/P)$ and $P$ is the size of a patch. Then each patch is flattened and linearly projected into a latent space by the image patch embedding layer. An extra learnable embedding is prepended as a token for classification (called "CLS") to the patch embeddings. Besides, position embeddings are added to the patch embeddings to retain positional information. In the second step, the resulting patch-embedding sequence, represented by ${\mathbf{E}}^i \in {\mathbb{R}^{(H{\times}W+1)\times{D}}}$ and $D$ denotes dimension of embeddings, is fed to the ViT encoder. A ViT encoder consists of $N^e$ stacked ViT encoder blocks, and each block contains a Multi-Head Self-Attention (MHSA) layer and a Position-wise Feed-Forward Network (PFFN). Both MHSA and PFFN have a layer normalization and a residual connection  applied to them. The output of ViT encoder is image feature ${\mathbf{F}}^i \in {\mathbb{R}^{(H{\times}W+1)\times{D}}}$, which consists of a sequence of patch features and a global classification feature. 

Then, we concatenate the features of all images in ${\mathbf{I}}^t$ and use extra $N^s$ stacked transformer encoder blocks \cite{vaswani2017attention} to establish the relationship between features from different images. The output of these extra encoder blocks is sequence feature ${\mathbf{F}}^e \in {\mathbb{R}^{(H{\times}W+1){K^t}\times{D}}}$, which is also the output of MUIsC encoder.

\subsubsection{Decoder}
The decoder is an autoregressive Natural Language Generation (NLG) model with vision-language fusion and an additional classification head. Our decoder is made up of a stack of $N^d$ decoder blocks, and each decoder block consists of three layers: a masked MHSA, a Multi-Head Cross-Attention (MHCA) and a PFFN. Each of these three layers is followed by a residual connection and a layer normalization. Masked MHSA has a similar architecture with the MHSA in MUIsC encoder. But different from MHSA that is applied to all tokens (i.e. bi-directional attention mechanism), masked MHSA only collects information from the prior tokens (i.e. uni-directional attention mechanism). MHCA is also similar with MHSA except that MHCA takes two sequences of embeddings/features as inputs, and then computes the relationship between them. In our method, MHCA's two inputs include the output from MUIsC encoder, i.e. ${\mathbf{F}}^e$ and the embedding sequence from the previous masked MHSA. Note that MHCA may be absent in some decoder blocks to keep our model concise. The input of the masked MHSA of the first decoder block is a sequence of $K^d$ word embeddings, $\mathbf{E}^d = \{{\mathbf{e^d}}_1,  ,...,{\mathbf{e^d}}_{K^d}\}$, which are generated from a sequence of words, denoted as $\mathbf{W}^d$. In the training stage, $\mathbf{W}^d$ is constructed by concatenating product title $\mathbf{W}^t$ and textual feedback $\mathbf{W}^f$ separated by a special token $s$, i.e. $\mathbf{W}^{d} = \{\mathrm{w}^t_1, ..., \mathrm{w}^t_{K^T}, \textrm{s}, \mathrm{w}^f_1, ..., \mathrm{w}^f_{K^f}\}$, where $\mathbf{W}^t$ provides additional textual information for a product and $\mathbf{W}^f$ serves as the groundtruth in the training of autoregressive NLG model. In the inference stage, $\mathbf{W}^d$ only contains $\mathbf{W}^t$ and a following $s$. We tokenize $\mathbf{W}^{d}$ into a sequence of tokens and encode the resulting tokens to word embeddings ${\mathbf{E}}^d$ via a word embedding layer. Then, position embeddings are added to ${\mathbf{E}}^d$. Taking word embeddings ${\mathbf{E}}^d$ and image-sequence feature ${\mathbf{F}}^e$ as input, our decoder finally outputs a sequence of decoded hidden states $\mathbf{H}^d$.

\subsubsection{ Multi-Task Learning} \label{multi-task learning}
We combine two tasks in a single MUIsC model - NLG which generates a textual feedback, and sequence Multi-class Classification (McC) which classifies a sequence to a qualified one or a category of rule violations. 

Given an input word sequence which consists of product title $\mathbf{W}^t$ paired with review feedback $\mathbf{W}^f$ and a sequence of images $\mathbf{I}^t$, our NLG task is to estimate the conditional probability for each token in $\mathbf{W}^f$:
\begin{align} 
\mathbf{P}^{nlg}(\mathbf{W}^f | \mathbf{I}^t, \mathbf{W}^t) = \prod^{K^f}_{q=1}p^{nlg}(w^f_q|w^f_{<q}; \mathbf{I}^t; \mathbf{W}^t)
\end{align}

where $w^f_{<q}$ stands for all tokens prior to position $q$ (i.e. $w^f_{<q} = (w^f_1, ..., w^f_{q-1})$). In the training stage, $\mathbf{W}^t$ is located prior to $\mathbf{W}^f$, and thus the $p^{nlg}(w^f_q)$ is conditioned on the preceding tokens in $\mathbf{W}^f$, $\mathbf{I}^t$, and all tokes in $\mathbf{W}^t$.

Given the whole training set $\mathcal{N}=(\mathcal{W}^t, \mathcal{I}^t, \mathcal{W}^f)$, our NLG task can be trained by optimizing the loss function as follows:
\begin{align} 
\mathcal{L}^{nlg}(\mathcal{N})=-\sum_{(\mathbf{W}^f,\mathbf{I}^t,\mathbf{W}^t)\in{\mathcal{N}}}\log{\mathbf{P}^{nlg}}(\mathbf{W}^f | \mathbf{I}^t, \mathbf{W}^t).
\end{align}
If we set textual feedback of a qualified image-sequence to a constant word, e,g. "yes", i.e. $K^f = 1$ and $w^f_1 = \textrm{"yes"}$, $p^{nlg}(w^f_1)$ is actually the probability of $\mathbf{I}^t$ being qualified. 

Sequence multi-class classification task classifies a sequence of images $\mathbf{I}^t$ into $K^g$ classes $\mathbf{G}\in\{g_1, ..., g_{K^g}\}$, where $g_1$ is the class of being qualified and $\{g_2, ..., g_{K^g}\}$ represent the classes of being rejected due to various rule violations. A McC head is applied on the decoded hidden state of token $s$ which follows $\mathbf{W}^t$, and thus our McC task is conditioned on $\mathbf{W}^t$ and $\mathbf{I}^t$. We use a softmax classifier where the loss function is

\begin{align*} 
\mathcal{L}^{mcc}(\mathcal{N})=
\sum_{(\mathbf{W}^f, \mathbf{I}^t, \mathbf{W}^t)\in{\mathcal{N}}} (-\sum^{K^g}_{i=1}1(c(\mathbf{I}^t)=i)\log(p^{mcc}_i(\mathbf{I}^t)))
\end{align*}, where $1(\cdot)$ is the indicator function, $c(\mathbf{I}^t)$ is the class label of $\mathbf{I}^t$, and $p^{mcc}_i(\mathbf{I}^t)$ represents the probability of $\mathbf{I}^t$ belonging to class i. Especially, $p^{mcc}_1(\mathbf{I}^t)$ is the probability of $\mathbf{I}^t$ being qualified.  

Combining the natural language generation and multi-class classification, the loss function for MUIsC is
\begin{align}\label{total_loss} 
\mathcal{L}(\mathcal{N}) = \lambda^{nlg}\mathcal{L}^{nlg}(\mathcal{N}) + \lambda^{mcc}\mathcal{L}^{mcc}(\mathcal{N})
\end{align}, where $\lambda^{nlg}, \lambda^{mcc}>0$ are factors that balance the two loss functions. In the inference stage, our framework takes
$p^{mcc}_1(\mathbf{I}^t)$
as its output, i.e. $p^t=p^{mcc}_1(\mathbf{I}^t)$.

\begin{table}
\centering
\begin{tabular}{ | c | c | c | c | c |} 

  \hline
  & { \normalsize Qualified} & { \normalsize Single-image} & {\normalsize Image-pair} & {\normalsize Multi-image}\\ 
  \hline
  Prop. & 0.71 & 0.12 & 0.07 & 0.10\\ 
  \hline
\end{tabular}
\caption{Proportions of qualified samples, and samples rejected due to single-image, image-pair, and multi-image rule violations in AGPIS-data. }
\label{tab:dataset_distribution}
\end{table}

\newcommand{\tabincell}[2]{\begin{tabular}{@{}#1@{}}#2\end{tabular}}
\begin{table*}
\centering
\begin{tabular}{ c | c | c | c | c | c | c | c | c | c } 
    \hline
    \multirow{2}{2cm}{Architecture} & Backbone & Fusion & AUC & R@P=0.8 & R@P=0.85 & R@P=0.9 & AUC-single & AUC-pair & AUC-multi\\
     &  & mode &  &  &  &  &  &  & \\
    \hline
    \multirow{6}{2cm}{Single-Tower} & \multirow{2}{4em}{ResNet18} & early & 0.745 & 0.803 & 0.592 & 0.344 & 0.739 & 0.736 & 0.737\\
    \cline{3-10}
    &  & \tabincell{c}{late} & 0.755 & 0.834 & 0.616 & 0.369 & 0.780 & 0.728 & 0.737\\
    \cline{2-10}
    &  \multirow{2}{4em}{ResNet50} & early & 0.751 & 0.806 & 0.620 & 0.415 & 0.746 & 0.735 & 0.744\\
    \cline{3-10}
    & & late & 0.763 & 0.841 & 0.647 & 0.392 & 0.792 & 0.724 & 0.743\\
    \cline{2-10}
    
    &  \multirow{2}{4em}{ResNetV2-101(BiT)} & early & 0.765 & 0.843 & 0.639 & 0.444 & 0.763 & 0.775 & 0.752\\
    \cline{3-10}
    & & late & 0.767 & 0.840 & 0.667 & 0.442 & 0.788 & 0.726 & 0.749\\
    \cline{2-10}
    & \multirow{2}{2em}{ViT} & early & 0.763 & 0.849 & 0.631 & 0.412 & 0.749 & 0.784 & 0.751\\
    \cline{3-10}
    &  & \tabincell{c}{late} & 0.764 & 0.831 & 0.644 & 0.429 & 0.772 & 0.729 & 0.767\\
    \cline{2-10}
    & ViT\textsuperscript{*} & \tabincell{c}{late} & 0.758 & 0.830 & 0.634 & 0.382 & 0.762 & 0.733 & 0.752\\
    \hline
    Ours& ViT & \tabincell{c}{late} & 0.800 & 0.904 & 0.758 & 0.524 & 0.795 & 0.774 & 0.791\\
    \hline
  
\end{tabular}
\caption{Performance comparison. Baselines use different visual backbones and image fusion methods. Superscript * indicates that the model is trained on a binary-class task and the other models are trained on a multi-class task.}
\label{tab:baseline_comparison}
\end{table*}

\section{Offline Evaluation of MUIsC}
\subsection{Dataset and Metrics} \label{dataset}
The stage 1 of our proposed framework is firstly deployed to produce image sequences for JD.com's "Haohuo" channel. With months of data accumulation, we collect produced three-image sequences and corresponding textual feedback as our dataset for MUIsC training and evaluation. Besides, textual product title and candidate images are also collected from JD.com's main site for each product in the dataset. In this paper, we use a dataset collected within three months and call it as AGPIS-data. AGPIS-data contains over 700K samples from 39 product categories. Distribution details of this dataset  can be found in Table \ref{tab:dataset_distribution}. We can see that about 29\% samples in our dataset are rejected ones. Note that, because part of image files are not valid anymore when we collect them and we only keep the samples with valid images, the proportion of qualified samples in our dataset is not equivalent to the acceptance rate in real production. 

We consider the rule name appeared in a textual feedback as the class label of a sample for the learning of multi-class classification task. If there are more than one rules in a textual feedback, we just choose the first one. In our dataset, only 4.7\% rejected samples have more than one violated rules. There are mainly 43 image-relevant rules in our AGPIS-data, and our multi-class labels have 45 classes with one extra class for qualified samples and another class for the samples rejected by other image-relevant rules. We can also convert multi-class labels to binary-class ones by simply merging 44 classes of rejected samples into one class. AGPIS-data is randomly split into three subsets without any overlap in product SKUs -- training (80\%), validation(10\%), and testing(10\%). Besides, in order to evaluate the detection performance of different categories of rule violations,  we also build datasets AGPIS-data-single, AGPIS-data-pair, and AGPIS-data-multi for single-image rules, image-pair rules, and multi-image rules, respectively. Each of the above datasets contains the same number of randomly selected qualified samples and samples rejected by a specific category of rule violations. 

ROC AUC (AUC) and Recall@Precision (R@P) are used to evaluate the performance of models. AUC is defined as the area under the ROC curve and R@P is the recall value at a given precision. 

\vspace{-1mm}
\subsection{Implementation details}
MUIsC is trained in an end-to-end manner with the encoder initialized by a pretrained ViT model and the decoder initialized by a pretrained GPT2 \cite{radford2019language} model. We use a pretrained ViT-B16-224 ($N^e=12$), the base version of the ViT with ($P=16$) patch size and ($H{\times}W=224\times224$) input image size. This model is pretrained on ImageNet-21k and finetuned on ImageNet 2012. For decoder, we use a GPT2 model pretrained on CLUECorpus \cite{xu2020cluecorpus2020},  a large-scale Chinese Corpus. Our decoder only has $N^d=3$ blocks to keep the model concise and efficient, since our textual product title and review feedback are relatively simple. Embedding dimension in both encoder and decoder is $D=768$. The model is trained for 10 epochs using the AdamW optimizer, with batch size 64. We use 1.5e-4 as the initial learning rate in our experiments and real applications.

Other parameters that need to be set in the proposed method are
\begin{itemize}
\item Number of images in target image sequence is $K^t=3$, and number of candidate images is $K^c\approx7$.
\item Number of extra encoder blocks for image-feature interaction is $N^s=1$.
\item Balance factor in the loss function in Eq. \ref{total_loss} is $\lambda^{nlg}=0.1$ and $\lambda^{mcc}=1.0$.

\end{itemize}

\begin{table*}
\centering
\begin{tabular}{ c | c | c | c | c | c | c | c | c | c | c | c | c } 
\hline
Single & Hier. &Enc.- & McC  & NLG  & Text  & AUC & R@P=0.8 & R@P=0.85 & R@P=0.9 & AUC- & AUC- & AUC-\\
tower & & dec. & task & task & as input &  &  &  &  & single & pair & multi\\
\hline
\checkmark & & & \checkmark & & &   0.764 & 0.831 & 0.644 & 0.429 & 0.772 & 0.729 & 0.767 \\
\checkmark & \checkmark & & \checkmark & & & 0.778 & 0.856 & 0.701 & 0.456 & 0.781 & 0.755 & 0.762 \\
& \checkmark & \checkmark & \checkmark &  &  &  0.780 & 0.876 & 0.688 & 0.450 & 0.784 & 0.757 & 0.762\\
& \checkmark & \checkmark & \checkmark & \checkmark &  & 0.792 & 0.891 & 0.730 & 0.507 & 0.799 & 0.764 & 0.765\\
& \checkmark & \checkmark & \checkmark & \checkmark & \checkmark & 0.800 & 0.904 & 0.758 & 0.524 & 0.795 & 0.774 & 0.791\\
\hline

\end{tabular}
\caption{Ablation of components in MUIsC.}
\label{tab:ablation_study}
\end{table*}

\subsection{Baseline Methods}

Our MUIsC aims to solve the problem of binary image-sequence classification, which can be considered as an extension of single-image classification. Note that a specific issue in image-sequence classification is how to fuse the information of multiple images in a sequence. To validate MUIsC, we build baselines based on the classic single-image classification architecture, called single-tower, which consists of a visual backbone and a classification head. We experiment with two image-fusion methods (early fusion and late fusion) and 4 backbones (ResNet18\cite{he2016deep}, ResNet50\cite{he2016deep}, ResNetV2-101\cite{kolesnikov2020big}, and the ViT\cite{dosovitskiy2020image} used in MUIsC). The early fusion method concatenates all images of a sequence into a single image, while the late fusion method firstly extracts features for each image and then concatenates these features together. The classification head is a MLP classifier. Models are all trained on AGPIS-data.

\subsection{Performance Comparison} \label{baseline_comparison}
Table\ref{tab:baseline_comparison} shows the performance of our method and baselines on different datasets, where AUC, AUC-single, AUC-pair, and AUC-multi represent the AUC on AGPIS-data, AGPIS-data-single, AGPIS-data-pair, and AGPIS-data-multi,  respectively. Besides the model with  superscript *, all models are trained on a multi-classification task. We can observe
that the proposed MUIsC outperforms other methods in AUC on AGPIS-data and achieves the best results in terms of all the other evaluation metrics and datasets except for AUC on AGPIS-data-pair.

\begin{figure*}[t]
  \includegraphics[width=1.0\textwidth]{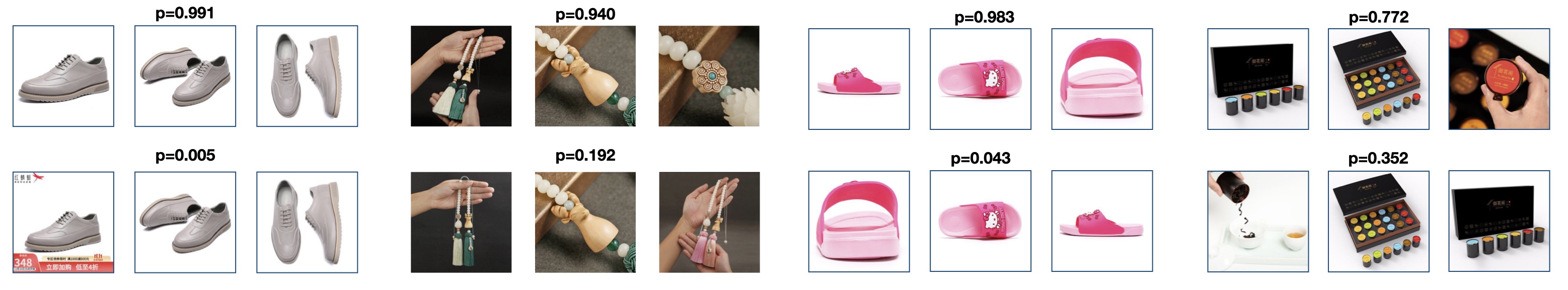}
  \caption{Examples of qualitative AGPIS framework results. }
  \Description{Examples of qualitative AGPIS results (image sequences with their probability of being qualified) by the proposed framework. The upper row show qualified sequences and the lower row show unqualified ones.}
  \label{fig:qualitative_images}
\end{figure*}

Among all visual backbones, ResNetV2-101 and ViT performs better than ResNet18 and ResNet50. This shows that powerful backbones are able to play some role in AGPIS. The comparison between early and late fusion modes indicates that each mode has its own strength and weakness, though late-fusion methods outperform early-fusion ones on the whole dataset. Early-fusion methods perform better on AGPIS-data-pair dataset. We think this is caused by the deep interaction between images. Late-fusion methods achieve better on AGPIS-data-single dataset since image features are extracted individually in the early stage. Our MUIsC adopts late-fusion considering the overall performance. But it is still an interesting topic to balance  model performance between single-image rules and image-pair rules. Besides, early-fusion and late-fusion methods have similar performance on AGPIS-data-multi. We speculate the reason is that AGPIS-data-multi emphasizes both single-image features and interaction between images. We also show the performance of the model trained on binary classification task, and observe that binary-class classifier performs worse than its corresponding multi-class one. This indicates that more detailed information about rejection can help improve model performance. In MUIsC, we further use textual feedback to 
introduce more information.

\vspace{0mm}
\subsection{Ablation Study}
The structure of MUIsC is carefully ablated with the results listed in Table \ref{tab:ablation_study}. Here, we use the same set of datasets and metrics with Section \ref{baseline_comparison}. The baseline for ablation study is a single-tower model trained on McC task, which either does not have decoder or uses any textual data. By introducing hierarchical image feature extraction method, a 1.4\% AUC gain is achieved, which shows the extra $N^s$ transformer encoder blocks result in better interaction between images than baseline late fusion. After using an encoder-decoder architecture, the performance is slightly improved, and indicates that directly adding a decoder 
does not lead to a big performance gain. Then, we
add the NLG task and formulate a multi-task learning model. It is observed that the performance is considerable improved, which indicates that the textual review feedback effectively guides the model to better understand rejected samples during training and results in better performance. By including textual product title as an extra input, we get a AUC gain of  0.8\% and the best performance is achieved. Interestingly, two kinds of textual information play different roles in terms of performance improvement. The NLG task on textual feedback leads to a good AUC gain on AGPIS-data-single, while product-title input achieves a significant gain on AGPIS-data-multi. This verifies that the detections of different categories of rule violations require different information for AGPIS.

\subsection{Qualitative Analysis}
Four qualitative example results are illustrated in Figure \ref{fig:qualitative_images}. Each example includes a "good" sequence and a "bad" one for a same product, according to the probability of being qualified $p^t$ estimated by our framework (shown above each sequence).  We can observe that all the sequences with low $p^t$ (the second row) violates our rules. The sequence in the leftmost example contains non-compliant banner and logo in the primary image. For the second left example, the primary and the third images contain products with different colors, which violates an image-pair rule. Besides, the second right and rightmost examples have improper display orders since the primary image fails to provide a whole picture of the product and violates multi-image rules. Meanwhile, we can see that all good sequences (the first row) have compliant single images and present the product in a proper order. These examples show that the proposed framework is effective in detecting all categories of rule violations and generating qualified image sequences.

\section{Online Evaluation}

\subsection{Deployment and Online Evaluation}
JD "Haohuo" Channel (Discovery Goods Channel) is an important traffic entrance for JD.com and also a good platform for users to discover their potential purchase interests, thus the product quality and presentation is of great influence for  platform's income. Before the release of our AGPIS framework, image selection for products mainly depends on human labors, which may be expensive and perform lower efficient. Since Feb 2021, our AGPIS framework has been deployed on "Haohuo" Channel (both website and App), and has produced high-standard images for about 1.5 million featured products. The amount of production is equivalent to 1000+ human professions.

Our framework is deployed step by step. Stage 1 is developed and deployed firstly. Then, stage 2 is trained on the resulting data and is deployed following stage 1. Our MUIsC model is updated every 3 months and is trained on AGPIS-data collected within the past 3 months. We take reject rate by human reviewers as online evaluation metric and set the submission threshold $th^t$ to 0.3. According to the statistical results, the reject rate is 19.3\% for the period when there is only stage 1, and it is further reduced to 13.6\% after stage 2 is added. Note that in order to avoid high reject rate in the period when stage 2 is not ready, stage 1 works under a strict setting, which may lead to high false-negative. In the future, we will gradually relax stage 1 and enable stage 2 to select among multiple candidates instead of just evaluating one candidate.

\subsection{System A/B testing}
In "Haohuo" Channel, an image-text system is built with our AGPIS framework and a product-copy generation framework\cite{guo2021intelligent}\cite{zhang2021automatic} to generate both images and textual description for a product. 
To evaluate the effectiveness of this system, we compare key online metrics before and after deploying our system on "Haohuo" Channel. The business of the platform is measured by Click-Through Rate (CTR) and Conversion Rate (CVR). The A/B testing is conducted on most of the main categories of products, covering clothes, electronics, computers, beauty \& health, groceries, etc. 
For users in baseline group, the platform shows them product images and text generated by human professions. 
For users in experiment group, the platform shows them images and text generated by our system. 
We find that images and text generated by our system outperform the ones submitted by human professions. Specifically, our system improves CTR by 3.2\% and CVR by 3.6\% over baseline. The increase of CTR indicates that the users prefer to click products with product images and text generated by our system. The improvement of CVR demonstrates that the images and text generated by our system is successful in convincing users to make purchase decisions.

Currently, the product image-text generation task in JD "Haohuo" Channel is jointly performed by human professions and our image-text system. We observe a trend that human professions sometimes prefer to use our system to assist their generation process. In addition, our system can also benefit for long-tail products, making the channel a healthier ecosystem.

\section{Conclusion}
In this paper, we developed a framework for Automatic Generation of Product-Image Sequence (AGPIS) in e-commerce. To address the unique challenges in AGPIS, our framework is designed as a combination of rule-based specific methods (stage 1) and a Multi-modality Unified Image-sequence Classifier (MUIsC) (stage 2) which is able to detect all categories of rule violations while considering multi-modality information. The experimental results show that our MUIsC outperforms various baselines. Our framework has been deployed on JD.com's "Haohuo" Channel, and a high volume of 1.5 million product image sequences have been generated. With the help of our AGPIS framework, our CTR is improved by 3.2\% and CVR is improved by 3.6\%.

\bibliographystyle{ACM-Reference-Format}
\bibliography{references}

\end{document}